\documentclass{article} 
\usepackage{iclr2020_conference,times}

\usepackage{amsmath,amsfonts,bm}

\def\eqref#1{equation~\ref{#1}}

\def\1{\bm{1}}

\DeclareMathAlphabet{\mathsfit}{\encodingdefault}{\sfdefault}{m}{sl}
\SetMathAlphabet{\mathsfit}{bold}{\encodingdefault}{\sfdefault}{bx}{n}

\usepackage{hyperref}
\usepackage{url}

\usepackage{float}
\usepackage[bf]{caption}    
\usepackage{subcaption}
\usepackage[linesnumbered,ruled]{algorithm2e}
\usepackage{amsmath}
\usepackage{graphicx}       

\def\a{\textit{a}}
\def\b{\textit{b}}

\title{How to augment your ViTs? Consistency loss and StyleAug, a random style transfer augmentation}

\author{Akash Umakantha \thanks{Work done during an internship at Bosch Center for Artificial Intelligence} \\
Carnegie Mellon University\\
Pittsburgh, PA 15213 \\
\texttt{aumakant@cmu.edu} \\
\And
Jo$\Tilde{\text{a}}$o D. Semedo\\
Bosch Center for Artificial Intelligence \\
Pittsburgh, PA 15222 \\
\texttt{joao.semedo@us.bosch.com} \\
\And
S. Alireza Golestaneh \\
Bosch Center for Artificial Intelligence \\
Pittsburgh, PA 15222 \\
\texttt{alireza.golestaneh@us.bosch.com} \\
\And
Wan-Yi S. Lin \\
Bosch Center for Artificial Intelligence \\
Pittsburgh, PA 15222 \\
\texttt{wan-yi.lin@us.bosch.com} \\
}

\iclrfinalcopy 
\begin{document}

\maketitle

\begin{abstract}
The Vision Transformer (ViT) architecture has recently achieved competitive performance across a variety of computer vision tasks. One of the motivations behind ViTs is weaker inductive biases, when compared to convolutional neural networks (CNNs). However this also makes ViTs more difficult to train. They require very large training datasets, heavy regularization, and strong data augmentations. The data augmentation strategies used to train ViTs have largely been inherited from CNN training, despite the significant differences between the two architectures. In this work, we empirical evaluated how different data augmentation strategies performed on CNN (e.g., ResNet) versus ViT architectures for image classification. We introduced a style transfer data augmentation, termed StyleAug, which worked best for training ViTs, while RandAugment and Augmix typically worked best for training CNNs. We also found that, in addition to a classification loss, using a consistency loss between multiple augmentations of the same image was especially helpful when training ViTs.
\end{abstract}

\section{Introduction}

For nearly a decade, convolutional neural networks (CNNs) have been the de-facto deep learning architecture for a variety of computer vision tasks from image classification to object detection to segmentation \citep{krizhevsky_imagenet_2012,redmon_you_2016,he_mask_2018,kolesnikov_big_2020}. A major reason for their success is the inductive biases imposed by the convolution operation, namely sparse interactions, weight sharing, and translational equivariance \citep{Goodfellow-et-al-2016}. These inductive biases allow for efficient training of feature representations that are useful for vision tasks.

Taking inspiration from the success of the Transformer architecture in language modeling \citep{vaswani_attention_2017}, Vision Transformers (ViTs) are alternative architectures that have recently shown promise for image classification, sometimes outperforming state-of-the-art CNNs \citep{dosovitskiy_image_2021, touvron_training_2021}. Follow-up work has shown that ViTs also have interesting properties and confer some advantages over CNNs, including: 1) increased adversarial and distribution-shift robustness \citep{shao_adversarial_2021,naseer_intriguing_2021}, 2) ability to provide pixel-level segmentation using attention maps \citep{caron_emerging_2021,naseer_intriguing_2021}, and 3) smaller texture bias and greater shape bias \citep{naseer_intriguing_2021,tuli_are_2021}. 

Although ViTs have attained competitive performance on vision tasks, they are known to be more difficult to train than CNNs \citep{steiner_how_2021, chen_when_2021, touvron_training_2021}. In ViTs, only multi-layer perceptron (MLP) layers operate locally and are translationally equivariant, while the self-attention layers \citep{vaswani_attention_2017} operate globally \citep{dosovitskiy_image_2021}. As such, ViTs are thought to have weaker inductive biases than CNNs, thus requiring more data, augmentations, and/or regularization than training similarly-sized CNNs \citep{chen_when_2021, steiner_how_2021, touvron_training_2021}. However, the strategies for data augmentation for training ViTs have largely been adapted from the techniques used for CNNs. While these augmentations have worked reasonably well, it is not clear whether the augmentations that work best for CNN training also work best for ViT training.

In this work, we performed a systematic empirical evaluation of data augmentation strategies for CNNs and ViTs. We found that the augmentations that worked best for ViTs were different from those that were best for CNNs. Furthermore, Using a consistency loss between different augmentations of the same image \citep{hendrycks_augmix_2020} was helpful when training ViTs, but not necessarily for training CNNs. We also introduced a style transfer data augmentation, termed StyleAug, inspired by shape bias in human visual perception \citep{geirhos_imagenet-trained_2018}. StyleAug performs neural style transfer from a given image to another randomly chosen image in the mini-batch during training. When combined with a consistency loss, StyleAug improved validation accuracy, robustness to corruptions, shape bias, and transferability to a different domain. For training ViTs, StyleAug outperformed previous state-of-the-art augmentations such as RandAugment \citep{cubuk_randaugment_2019} and AugMix \citep{hendrycks_augmix_2020} across several metrics.

\section{Related work}

\textbf{ViT training.} ViTs have weaker inductive biases than CNNs \citep{steiner_how_2021, chen_when_2021, touvron_training_2021}. To achieve classification performance better than CNNs, \citet{dosovitskiy_image_2021}, trained ViTs on very large datasets, either ImageNet-21k or the proprietary JFT-300M. To train ViTs with limited data and compute resources, \citet{steiner_how_2021} explored data augmentations and regularization. They suggested that, for a fixed dataset size, one should generally prefer data augmentations over extensive regularization. \citet{touvron_training_2021} trained data-efficient vision transformers using a combination of various augmentations, regularization strategies, and a novel distillation strategy. For distillation, they created a special "distillation token" in the transformer architecture that uses a CNN as the teacher network. Their data-efficient image transformer (DeiT) achieved competitive performance without large datasets (e.g., ImageNet-1k).

\textbf{Data augmentation.} Data augmentation can increase the size and quality of datasets, which can help prevent overfitting and greatly improve generalization of deep learning models. Since ViTs have weaker inductive biases, they can be prone to overfitting \citep{chen_when_2021}, and thus benefit greatly from many strong augmentations \citep{touvron_training_2021}. 

From another perspective, data augmentation can also help deep learning models learn invariances such as scale (i.e., with cropping) and color. Indeed, increasingly popular self-supervised learning methods learn feature representations by becoming invariant to image transformations. \citet{caron_emerging_2021} showed that multi-scale cropping is an especially useful augmentation for training self-supervised ViTs. \citet{hendrycks_augmix_2020} took inspiration from the self-supervised learning literature, and used a Jensen-Shannon consistency loss (between a training image, and multiple augmentations of the image) in addition to a classification loss when training CNNs.

\textbf{Shape vs. texture bias} \citet{geirhos_imagenet-trained_2018} used psychophysics experiments to show that humans make image classification decisions based on object shape, rather than relying on image texture. Presented with the same images, CNNs made decisions more biased on image texture. \citet{geirhos_imagenet-trained_2018} created a new dataset, called Stylized ImageNet, in which they performed style transfer with ImageNet as content images, and art as style images. Trained on this data, CNNs showed improved shape bias and lower texture bias, but at the expense of classification performance. On the other hand, \citet{naseer_intriguing_2021} showed that ViTs tend to have higher shape bias than CNNs. 

\citet{xu_robust_2021} used a random convolution augmentation (to distort textures) combined with a consistency loss to improve CNN generalization to unseen domains such as ImageNet-sketch. \citet{jackson_style_2019} used style transfer to paintings (Painter by Numbers) as an augmentation technique. When combined with traditional augmentations, their augmentation improved CNN classification and domain transfer performance on several small datasets, and on monocular depth estimation in the KITTI dataset.


\begin{figure}[t]
    \captionsetup{labelformat=simple}
    \begin{center}
    \includegraphics[width=1\textwidth]{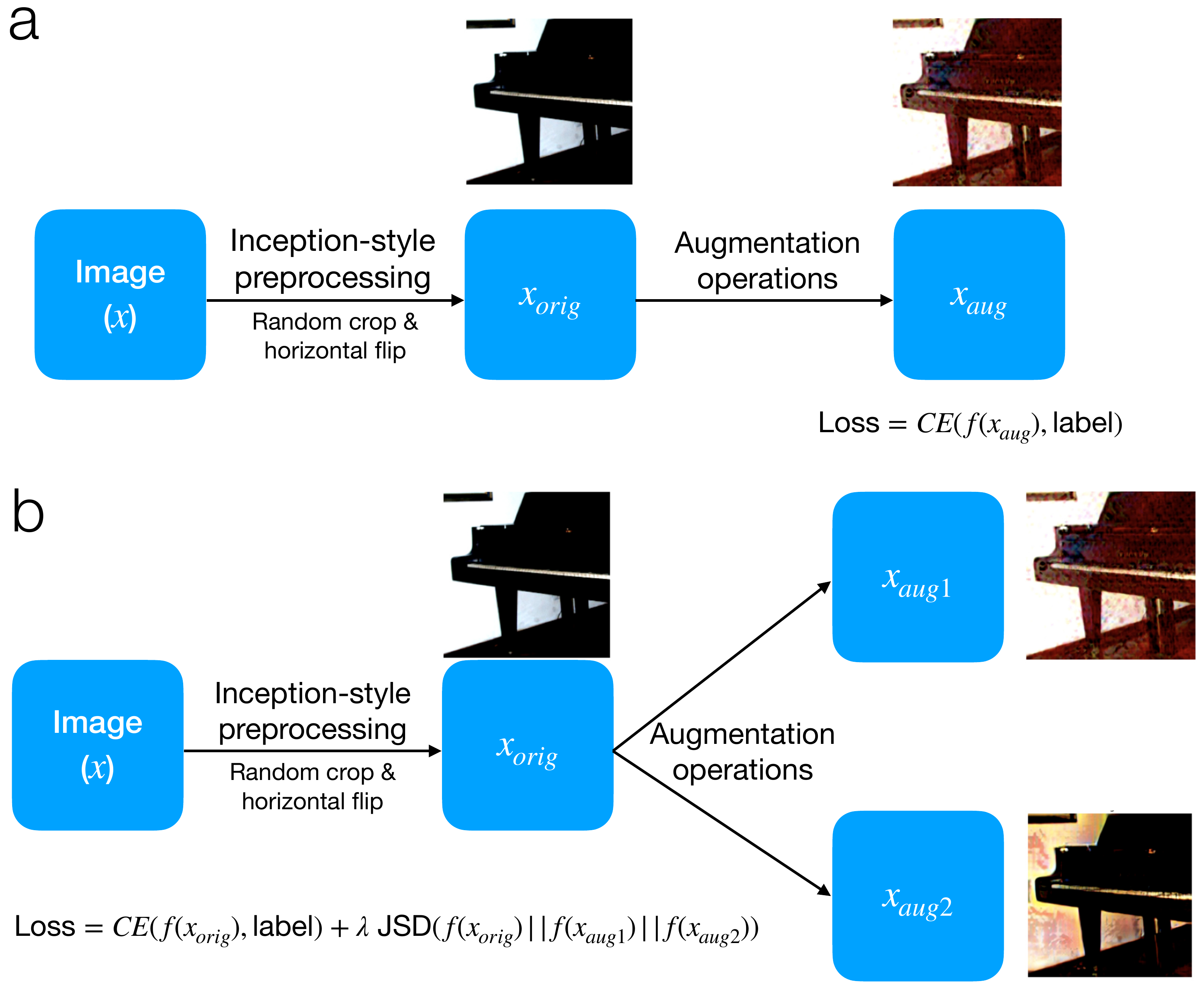}
    \caption[Augmentation setup and JSD consistency loss function.]{\textbf{Augmentation setup.} (a) Classic augmentation setup. Cross-entropy loss between the network prediction of the augmented image, $ f(x_{aug1}) $, and the true label. (b) Setup with a Jensen-Shannon (JSD) consistency loss. Cross-entropy loss between prediction of the original image $ f(x_{orig}) $, and an addition of a JSD loss between the three network predictions of each of the original image ($ f(x_{orig}) $) and two augmentations ($ f(x_{aug1}) $, $ f(x_{aug2}) $).}
    \label{fig:aug_setup}
    \end{center}
\end{figure}

\section{Augmentation strategies}

\subsection{Image transformations}

For training models, we tested both basic and state-of-the-art augmentation strategies for image classification. All images first went through Inception-style preprocessing: 1) a resized crop with a randomly chosen scale in $[0.5,1]$ and resized to $224\times224$, and 2) a random horizontal flip with $p=0.5$ (Fig. \ref{fig:aug_setup}\a, ImageNet Image to "Orig"). We used a relatively large cropping scale in this step so allow for testing of multi-scale cropping augmentations later (see JSD loss below; \citep{caron_emerging_2021}). 

We then performed additional augmentation operations to this image (Fig. \ref{fig:aug_setup}\a, "Orig" to "Aug 1"). First, we tested basic augmentations such as random cropping, color jittering, and translation. Second, we tested RandAugment \citep{cubuk_randaugment_2019}, a state-of-the-art augmentation for training CNNs on ImageNet, and AugMix \citep{hendrycks_augmix_2020}, another state-of-the-art augmentation that improves CNN robustness to image corruptions. As in the Augmix paper, for both RandAugment and AugMix, we exclude transformations that overlap with ImageNet-C corruptions to allow for fair evaluations of model generalization and robustness. Third, we tested our new human perception-inspired augmentation StyleAug (described in detail below), and StyleAug with random cropping. Finally, we also tested another augmentation, called Neurofovea (Deza \textit{et al.}, 2021 \citep{deza_emergent_2021}), inspired by the human perceptual phenomena of foveation and metamerism \citep{freeman_metamers_2011}.

In experiments, we considered the random cropping augmentation as a baseline as it is effectively the same as only training models with Inception-style preprocessing. Examples of all augmentations tested, along with further details such as any torchvision transforms used, are provided in the Appendix and Supp. Fig. \ref{appendix_fig:aug_examples}. 

\subsection{Jensen-Shannon divergence (JSD) consistency loss}

For the typical training augmentation setup (Fig. \ref{fig:aug_setup}\a), we trained models using a cross-entropy classification loss (with label smoothing=0.1) between the model predictions (i.e., $f(\hat{y}|x_{aug1})$ posterior distribution over class labels given an image "Aug 1") and the true class label $y$. We tested the impact of using a consistency loss to train different model architectures. Following the AugMix paper \citep{hendrycks_augmix_2020}, we used a Jensen-Shannon divergence (JSD) consistency loss between an image ($x_{orig}$) and two augmentations of the image ($x_{aug1}$ and $x_{aug2}$). This JSD loss was applied in addition to a classification loss (Fig. \ref{fig:aug_setup}\b):
\begin{align} \label{eqn:loss}
    \mathcal{L}(f(\hat{y} | x_{orig}) , y)+\lambda\text{JSD}(f(\hat{y} | x_{orig}) || f(\hat{y} | x_{aug1}) || f(\hat{y} | x_{aug2}))  
\end{align}

We used $\lambda=12$, the value used in AugMix \citep{hendrycks_augmix_2020}. The JSD loss is computed as follows:
\begin{align} \label{eqn:jsd}
    \text{JSD}(p_{orig} || p_{aug1} || p_{aug2}) = \frac{1}{3} \left( KL(p_{orig} || M ) + KL(p_{aug1} || M ) + KL(p_{aug2} || M ) \right)
\end{align}

where $KL$ is the KL divergence, and $M = (p_{orig} + p_{aug1} + p_{aug2})/3$.
The JSD loss imposes a large penalty when the posterior distribution predictions for the three versions of the training image are very different. Thus, the JSD consistency loss requires models to learn similar feature representations and output distributions across the different augmented versions of the same image. This explicitly trains models to become invariant to the augmentations used.


\section{StyleAug}

\begin{figure}[b]
    \captionsetup{labelformat=simple}
    \begin{center}
    \includegraphics[width=1.0\textwidth]{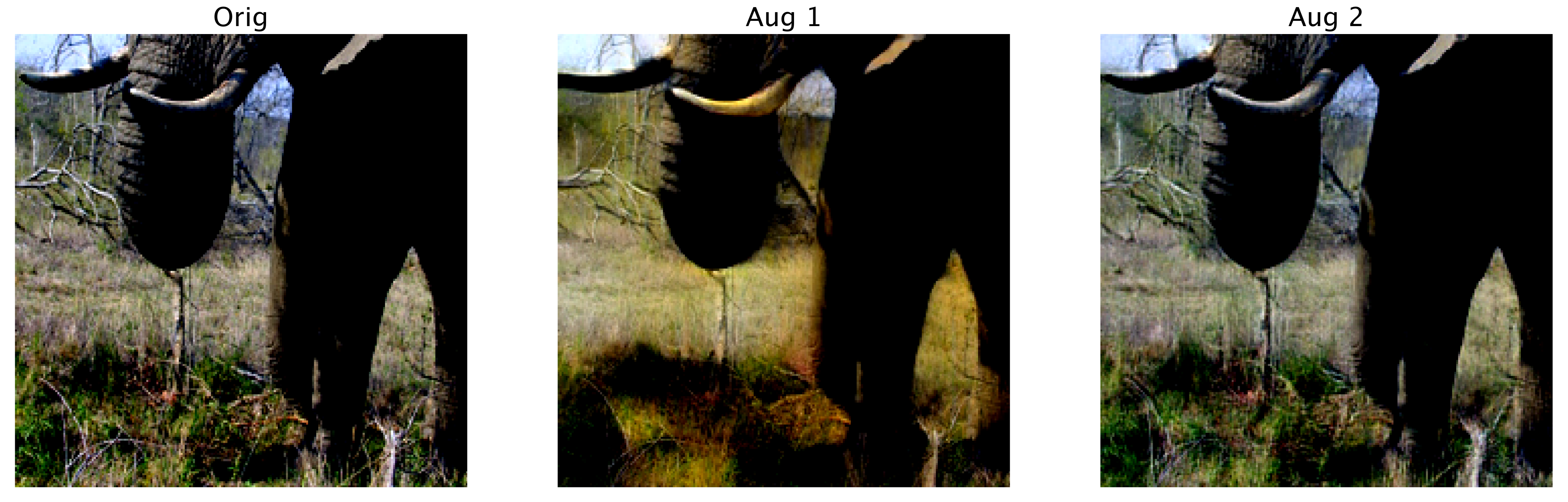}
    \caption[StyleAug augmentation example]{\textbf{StyleAug:} neural style transfer from a given image in the batch to another randomly chosen image in the current mini-batch. "Orig" (left) shows the original image after Inception-style preprocessing; "Aug 1" (middle) and "Aug 2" (right) show two StyleAug augmentations of "Orig".}
    \label{fig:styleAug_images}
    \end{center}
\end{figure}

\citet{geirhos_imagenet-trained_2018} showed that CNNs trained on ImageNet make classification decision mainly based on image textures (i.e., they have high texture bias). However, high shape bias and low texture bias might be desirable because models with this property tend to show better generalization and increased robustness \citep{xu_robust_2021,naseer_intriguing_2021,geirhos_imagenet-trained_2018}. New datasets (i.e., Stylized ImageNet \citep{geirhos_imagenet-trained_2018}) and augmentations (i.e., random convolutions \citep{xu_robust_2021}) have been developed to try to improve CNN shape bias, robustness, and generalization. However, training with these techniques is expensive and/or does not improve validation accuracy on the original ImageNet dataset. We sought to develop an augmentation that: 1) is fast and can be used in real-time during training, 2) improves shape bias, and 3) improves performance on ImageNet.

We introduce a data augmentation termed StyleAug (Fig. \ref{fig:styleAug_images}). StyleAug combines aspects of the style transfer in Stylized ImageNet \citep{geirhos_imagenet-trained_2018} as well as the JSD consistency loss used in the random convolution augmentation \citep{xu_robust_2021} and AugMix \citep{hendrycks_augmix_2020}. StyleAug uses adaptive instance normalization (AdaIn; \citep{huang_arbitrary_2017}) for fast real-time arbitrary style transfer. We use a training image (whose label is preserved) as the content image, and another randomly chosen image in the batch as the style image. To ensure that the training label is preserved, the augmented image is a mix of the original image and the style-transferred image, where the mixing weight $m$ was drawn from a $\beta(50,50)$ distribution (i.e., most of the time $m$ was close to 0.5, but there was some stochasticity in the amount of style distortion). In terms of computational resources, training models with StyleAug used approximately the same amount of time and resources as training models with RandAugment or AugMix. 

StyleAug tended to preserve the shape content of an image but distorted its colors and textures (e.g., Fig. \ref{fig:styleAug_images}). By combining StyleAug with the JSD consistency loss, we explicitly trained networks to become invariant to the color, texture, and other distortions/transformations that were induced by StyleAug. Pseudocode for StyleAug training with the JSD loss is provided in Algorithm \ref{algo:styleaug}.

StyleAug differs from the work of \citet{jackson_style_2019} in several ways. First, instead of performing style transfer to a different dataset (e.g., art or painting datasets), we performed style transfer to randomly chosen images within the \textit{same} mini-batch (i.e., the same dataset). Second, we combined StyleAug with a Jensen-Shannon divergence consistency loss, which we found typically improves performance, especially for ViTs. We also used a different fast arbitrary style transfer methodology (AdaIn). Finally, we evaluated StyleAug on ImageNet as well as evaluated robustness to image corruptions, shape bias, and transfer learning performance.

\begin{algorithm}[t]
    \SetKwProg{Fn}{Function}{:}{}
    \DontPrintSemicolon
    \SetKwInOut{Input}{Input}
    \SetKwInOut{Output}{Loss Output}
    \SetKwFunction{StyleAug}{StyleAug}
    \Input{Model $f$, classification loss $\mathcal{L}$, training image $x$ and its class label $y$, two images sampled randomly from the current mini-batch $x_{rand1},x_{rand2}$}
    \;
    \Fn{\StyleAug{$x$,$x_{style}$,$\alpha=50$,$\beta=50$}}{
        $z = VGG_{enc}(x)$ \tcp*{VGG encoder from \citet{huang_arbitrary_2017}}
        $z_{style} = VGG_{enc}(x_{style})$\;
        $z_{adain} = AdaIn(z,z_{style})$ \tcp*{adaptive instance normalization}
        $x_{adain} = VGG_{dec}(z_{adain})$ \tcp*{VGG decoder}
        $m\sim Beta(\alpha,\beta)$\;
        $x_{aug} = m\cdot x + (1-m)\cdot x_{adain}$ \tcp*{mix with original image}
        \KwRet $x_{aug}$\;
    }
    \;
    $x_{orig} = \text{InceptionStylePreprocess}(x)$ \tcp*{Random crop and horizontal flip}
    $x_{style1} = \text{InceptionStylePreprocess}(x_{rand1})$\;
    $x_{style2} = \text{InceptionStylePreprocess}(x_{rand2})$\;
    \;
    $x_{aug1} = \StyleAug(x_{orig},x_{style1})$\;
    $x_{aug2} = \StyleAug(x_{orig},x_{style2})$ \tcp*{$x_{aug1} \neq x_{aug2}$}\;
    
    \Output{$\mathcal{L}(f(\hat{y} | x_{orig}) , y)+\lambda\text{JSD}(f(\hat{y} | x_{orig}) || f(\hat{y} | x_{aug1}) || f(\hat{y} | x_{aug2}))$ }
    \caption{StyleAug training with Jensen-Shannon (JSD) consistency loss}
    \label{algo:styleaug}
\end{algorithm}


\section{Experiments}

For fair comparison, we trained models of similar size: ResNet-50 for CNNs ($\sim 25$ million parameters), and ViT-Small with 16x16 patch size for vision transformer ($\sim 22$ million parameters) \citep{chen_when_2021}. All models were trained from random initialization for 100 epochs on ImageNet-1k. We used the AdamW optimizer with a peak learning rate of 0.001 with linear warmup for 10 epochs followed by a cosine learning rate decay schedule. For ResNet-50 training, we used a weight decay of 0.05, a batch size of 512 for typical training (no JSD) and 200 for training with the JSD consistency loss (since the JSD loss uses $\sim3\times$ the number of images per batch). ViT-Small/16 required more GPU memory during training, requiring smaller batch sizes of 400 (no JSD) and 150 (JSD). We also used a larger weight decay of 0.3 for ViT-Small/16. We trained models using 2 GPUs (Nvidia Tesla V100) in parallel for training without the JSD loss, and 4 GPUs for training with the JSD loss. Basic augmentations (crop, color, translate) required approximately 30-45 minutes per epoch for training, while the other augmentations required approximately 60-75 minutes.

We trained models using the augmentations described in Sections 3 and 4, with and without a JSD consistency loss, and evaluated their performance on:
\begin{enumerate}
    \item ImageNet validation accuracy
    \item Robustness to corruptions / distribution shift (i.e., accuracy on ImageNet-C \citep{hendrycks_benchmarking_2019}) 
    \item Shape bias vs. texture bias on cue-conflict images \citep{geirhos_imagenet-trained_2018}
    \item Transfer learning to: The Oxford-IIIT Pet dataset (pet37, which is a dataset of natural images \citep{parkhi_cats_2012}) and to resisc45 (a dataset of satellite images \citep{cheng_remote_2017}).
\end{enumerate}

We had two key findings. First, we found that using a JSD consistency loss was almost always beneficial for ViTs, while using the consistency loss provided mixed results for CNNs. Second, for both CNNs and ViTs, StyleAug greatly improved shape bias over other augmentations we tested (Fig. \ref{fig:shape_bias}). Moreover, for ViTs, StyleAug also provided the best ImageNet validation performance (Fig. \ref{fig:imagenet_1k}), mean corruption accuracy on ImageNet-C (Fig. \ref{fig:imagenet_c}, and transfer learning performance to the Pet37 dataset (Fig. \ref{fig:transfer}\a).

\subsection{ImageNet-1k validation accuracy}

\begin{figure}[!b]
    \captionsetup{labelformat=simple}
    \begin{center}
    \includegraphics[width=0.9\textwidth]{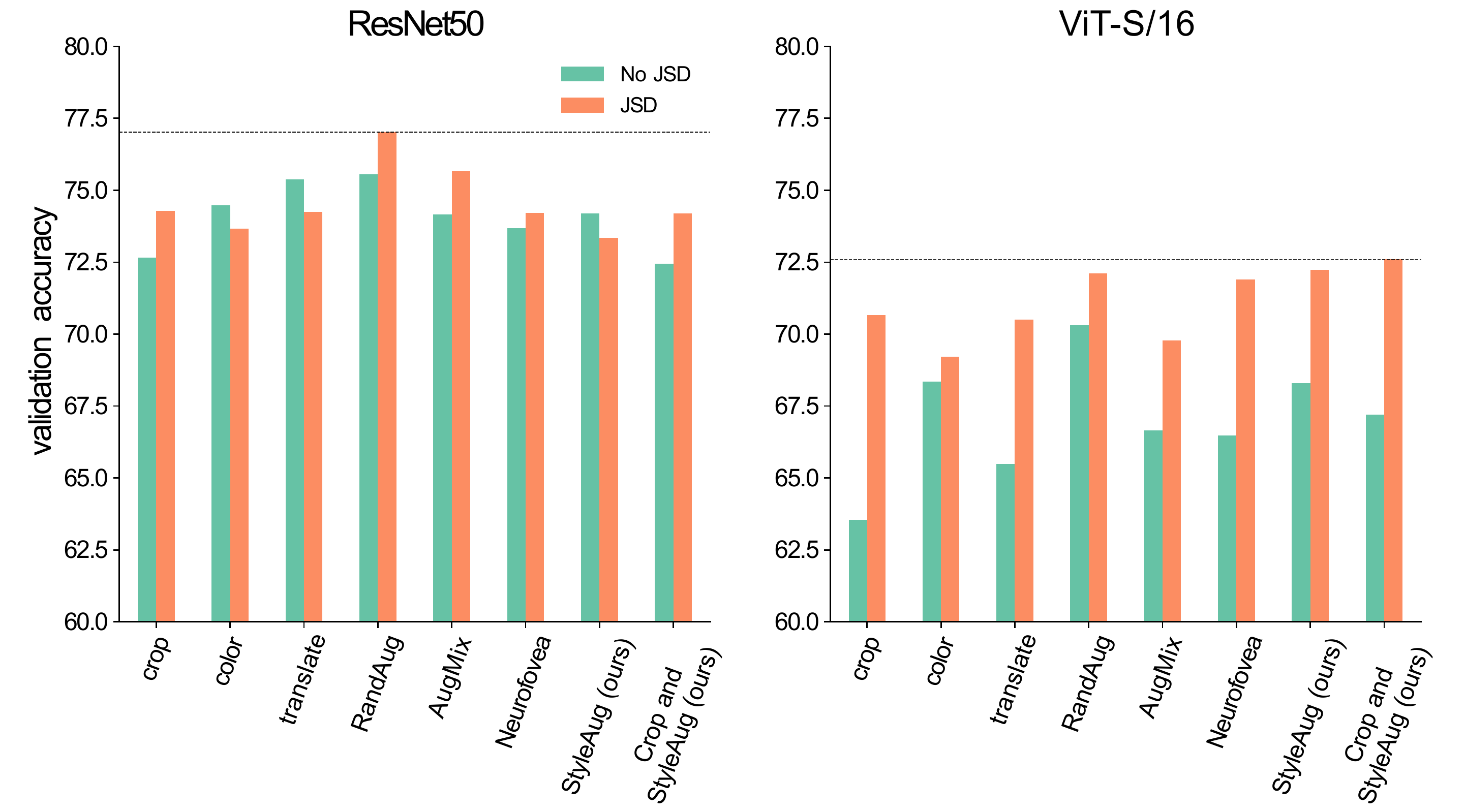}
    \caption[ImageNet-1k validation accuracy.]{\textbf{Validation accuracy of different augmentations on ImageNet-1k.}}
    \label{fig:imagenet_1k}
    \end{center}
\end{figure}

To evaluate models' performance on the ImageNet-1k, we preprocessed validation images by resizing to 256 pixels, and then taking a $224\times224$ center crop. Here, we report the accuracy of models on the ImageNet-1k validation set. We note that ViT-S validation accuracy was lower than ResNet-50 performance. This is expected when training on the relatively small ImageNet-1k; ViT models that outperformed ResNets were trained on larger datasets (ImageNet-21k or JFT-300M \citep{dosovitskiy_image_2021}) or used knowledge distillation \citep{touvron_training_2021}. Here, we were most interested in comparing the relative performance improvements of using one augmentation strategy over another, and how that differed between CNNs and ViTs.

Most importantly, we found that using a JSD consistency loss provided a large boost in accuracy across all augmentations when training ViT-S (Fig. \ref{fig:imagenet_1k} right panel, orange bars all above green bars). For ResNet-50, using a JSD loss improved accuracy for some augmentations but resulted in lower accuracy for others (Fig. \ref{fig:imagenet_1k} left panel).

Second, we found that the augmentations that worked best for ResNet-50 were different from those that worked best for ViT-S. For ViT-S, our proposed augmentation, StyleAug and StyleAug + crop, have the best accuracy, followed closely by RandAugment and Neurofovea. For ResNet-50, the state-of-the-art RandAugment and Augmix perform best.

Finally, we note that our cropping + JSD loss augmentation (see Supp. Fig. \ref{appendix_fig:aug_examples}\a) is very similar to multi-scale cropping used in DINO \citep{caron_emerging_2021}. Thus, our finding that using a JSD consistency loss with random cropping supports the finding in the DINO paper that multi-scale cropping in a self-supervised setting is a very beneficial augmentations for ViTs.

\subsection{Robustness to corruptions}

We tested the models trained on ImageNet-1k on their robustness to distribution shift (i.e., image corruptions). To do so, we evaluated each models' performance on ImageNet-C \citep{hendrycks_benchmarking_2019}, which contains 19 different corruptions across 5 different severity levels each. We report the mean corruption accuracy as the model's average accuracy across the 95 datasets present in ImageNet-C.

\begin{figure}[!b]
    \captionsetup{labelformat=simple}
    \begin{center}
    \includegraphics[width=0.9\textwidth]{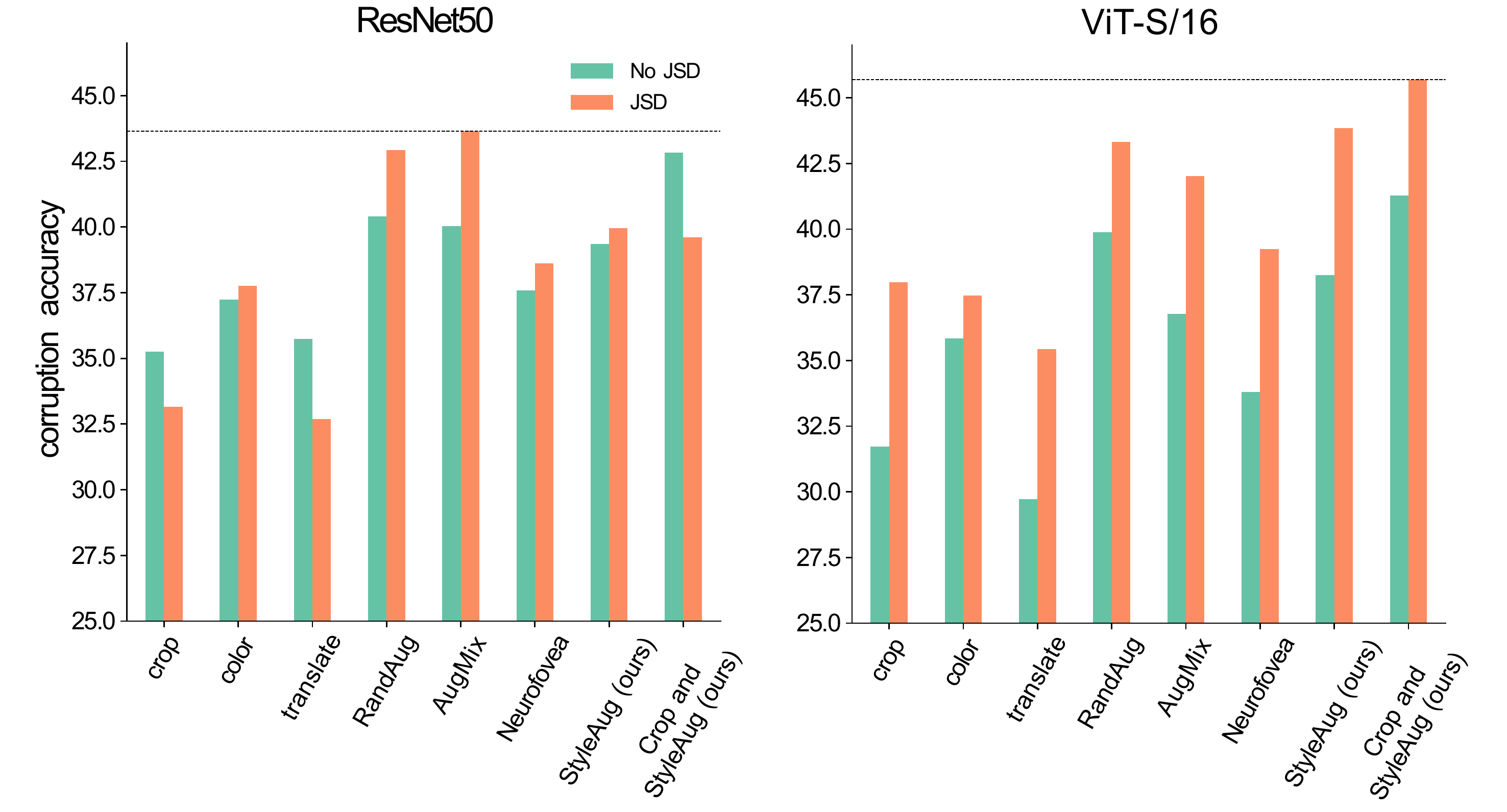}
    \caption[Robustness to distribution shift: ImageNet-C mean corruption accuracy]{\textbf{Mean corruption accuracy of different augmentation strategies on ImageNet-C.}}
    \label{fig:imagenet_c}
    \end{center}
\end{figure}

For ViT models, we found that training with StyleAug and a JSD consistency loss attained the highest corruption accuracy, and in fact also outperformed all ResNet-50 models (Fig. \ref{fig:imagenet_c}). RandAugment and Augmix with a JSD loss had the highest corruption accuracy among ResNet-50 models. Secondly, we again found that using the JSD consistency loss during training boosted the corruption accuracy of ViT models by close to 5\% in many cases (Fig. \ref{fig:imagenet_c} right panel, orange bars above green bars). However, using a JSD loss again provided mixed results in corruption accuracy for ResNet-50 models (Fig. \ref{fig:imagenet_c} left panel).

We also note that while ImageNet-1k validation accuracy was lower for ViT-S than ResNet-50, here we found that, for the same models as in Fig. \ref{fig:imagenet_1k}, many ViT-S models had higher corruption accuracy than ResNet-50 models. This supports findings in \citet{naseer_intriguing_2021} that suggest that vision transformers tend to be more robust than CNNs.

\subsection{Shape bias}

We tested each models' shape bias relative to its texture bias. To do so, we evaluate the models trained on ImageNet-1k on the cue-conflict images from Geirhos \textit{et al.} \citep{geirhos_imagenet-trained_2018}. Cue-conflict images were generated by performing iterative style transfer between two images. Thus, they have an object shape label (based on the content image) and an texture label (base on the style image; see Supp. Fig. \ref{appendix_fig:cueConflict} for an example). The shape bias is defined as the number of correctly classified shape labels relative to the total number of correctly classified images (either shape or texture) \citep{geirhos_imagenet-trained_2018}:
$$ \text{shape bias} = \frac{\# \text{ correct shape labels}}{\# \text{ correct shape labels} + \# \text{ correct texture labels}} $$

\begin{figure}[!t]
    \captionsetup{labelformat=simple}
    \begin{center}
    \includegraphics[width=0.9\textwidth]{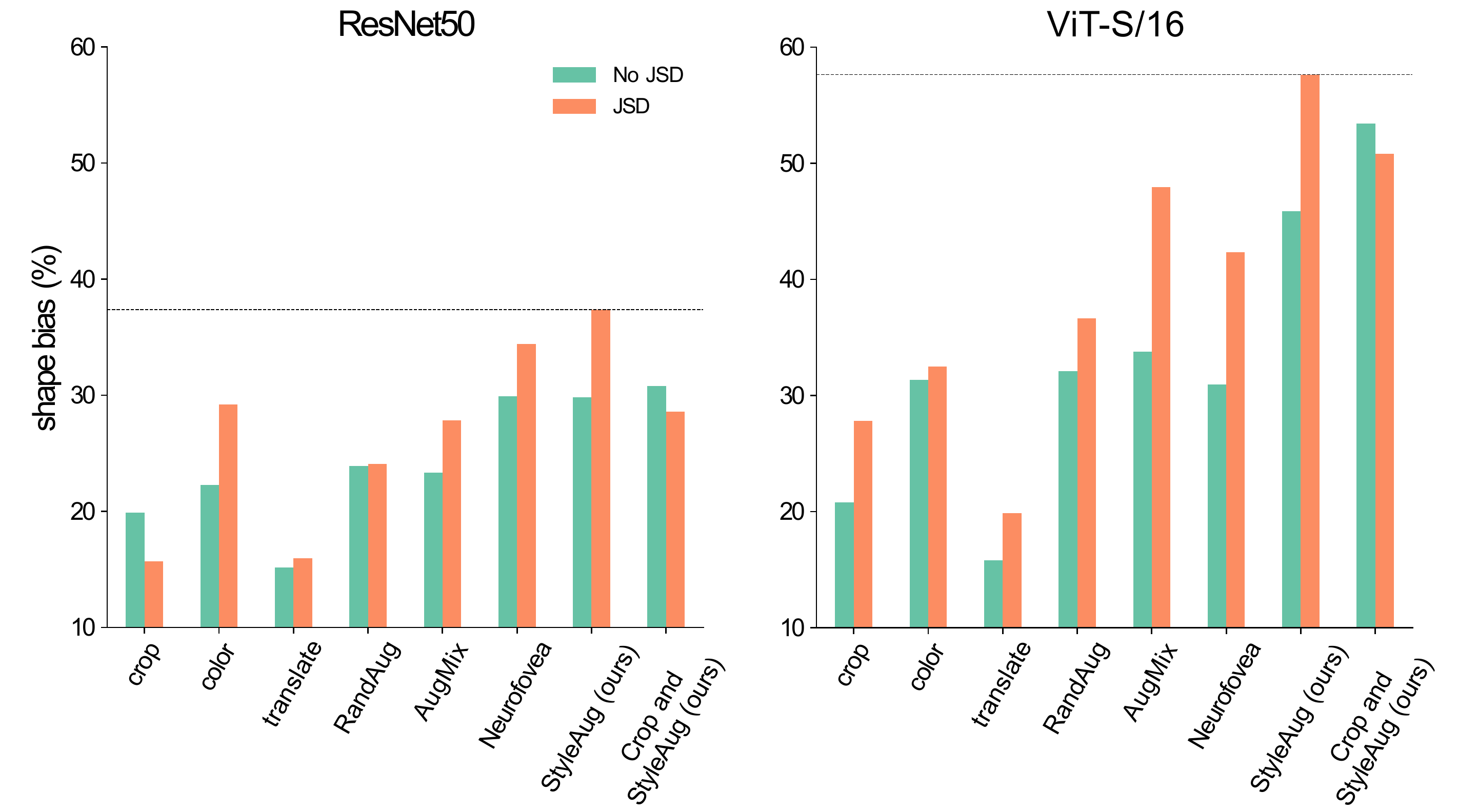}
    \caption[Shape bias of ImageNet trained models]{\textbf{Shape bias of different augmentation strategies.}}
    \label{fig:shape_bias}
    \end{center}
\end{figure}

We found that training ViT-S with the JSD consistency loss greatly improved shape bias (Fig. \ref{fig:shape_bias} right panel, orange bars above green bars), and that StyleAug provided the highest shape bias. Training with the JSD loss also tended to increase shape bias across augmentations for ResNet-50, and we also found that our proposed StyleAug provided the highest shape bias for ResNet-50 (Fig. \ref{fig:shape_bias} left panel). However, we note that ResNet-50 models had much lower shape bias than ViT-S models.

\subsection{Transfer learning}

We tested the transferability of models trained in on ImageNet-1k. To do so, we froze the backbone weights, replaced the classification heads, and only finetuned the weights of the new classification heads on Pet37 or Resisc45. For Pet37, we used SGD with momentum, with a batch size of 512, and learning rate of 0.01 for 10 epochs, followed by a learning rate of 0.003 for 10 epochs. We evaluated performance on the test split used in \citep{parkhi_cats_2012}. For Resisc45, we used SGD with momentum, with a batch size of 512, and learning rate of 0.01 for 10 epochs, 0.003 for 5 epochs, and 0.001 for 5 epochs. Since there is not a standard training / test split, we performed a single random 80/20 split which was kept constant across training and evaluation of different models. We did not use any augmentation during transfer learning; we only used augmentations while training on ImageNet-1k. 

\begin{figure}[!t]
    \captionsetup{labelformat=simple}
    \begin{center}
    \includegraphics[width=0.9\textwidth]{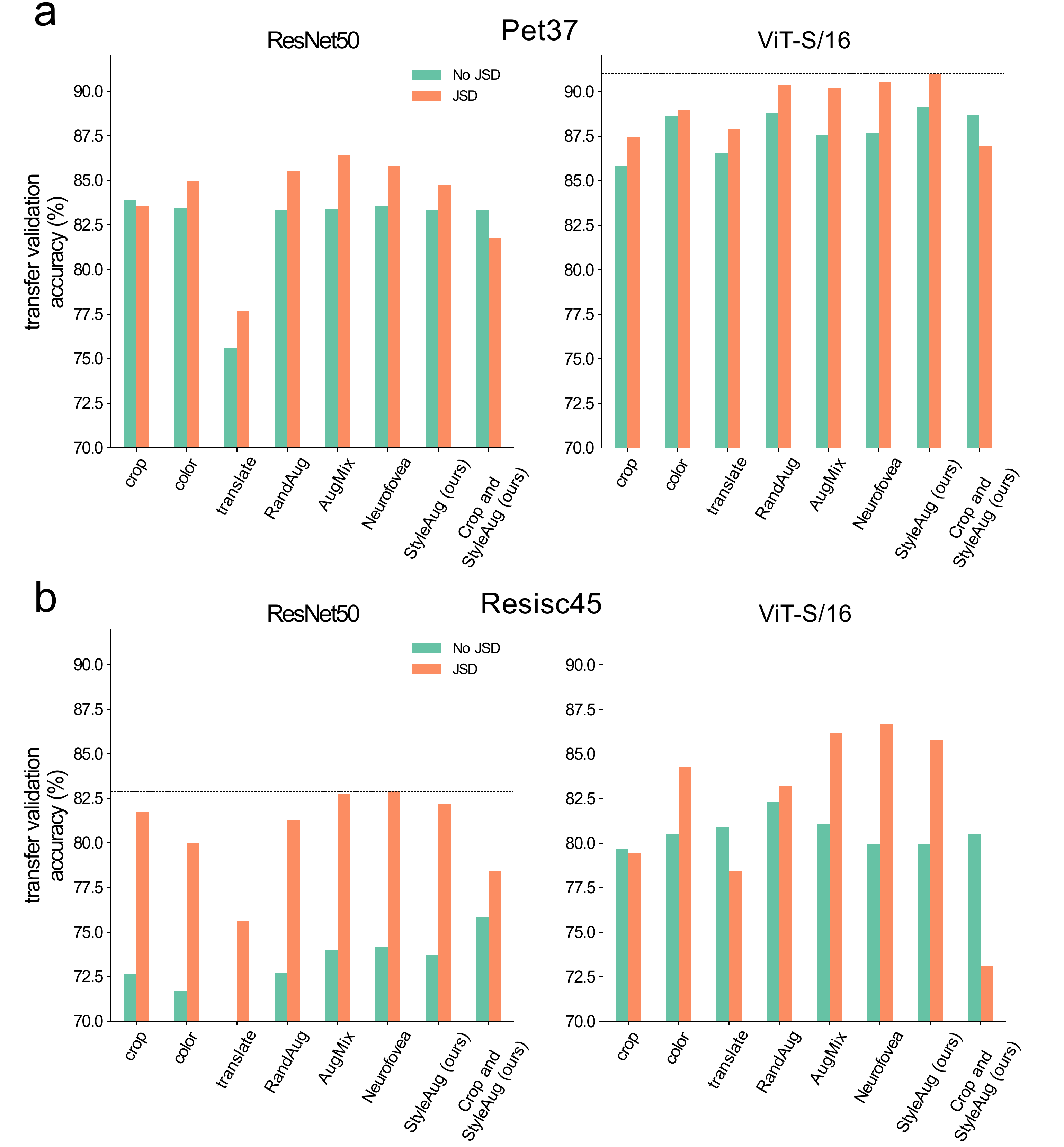}
    \caption[Transfer learning of ImageNet trained models to Pet37 and Resisc45.]{\textbf{Transfer learning of ImageNet trained models.} (a) Validation accuracy on Pet37 after transfer learning. (b) Validation accuracy on Resisc45 after transfer learning.}
    \label{fig:transfer}
    \end{center}
\end{figure}

For both Pet37 and Resisc45, we found that JSD consistency loss improved transfer learning for both ResNet-50 and ViT-S models (Fig. \ref{fig:transfer}\a-\b, orange bars typically above green bars). For both datasets, we found that ViT-S models transferred better than ResNet-50 models (Fig. \ref{fig:transfer}\a-\b, bars in right panels higher than those in left panels). For Pet37, StyleAug worked best again for ViT-S while AugMix worked best for ResNet-50 (Fig. \ref{fig:transfer}\a). For the satellite images of Resisic45, Neurofovea, Augmix, and StyleAug worked well for both ViT-S and ResNet-50 (Fig. \ref{fig:transfer}\b).


\section{Conclusion}

In this work, we systematically evaluated how different commonly-used augmentation strategies perform on different model architectures. We showed that the data augmentations that work best for ViTs are different than those that work best for CNNs. Furthermore, we found that using a Jensen-Shannon consistency loss in addition to a classification loss when training ViTs provided considerable performance improvement in almost all cases. Importantly, although ViT performance lagged CNN performance on ImageNet-1k, they were generally more robust to corruptions, had higher shape bias, and were more transferable. We hope that future work will scrutinize other existing data augmentations and training strategies that have worked well for CNNs, and consider whether they should be used for other model architectures like ViTs.

We also introduced StyleAug: real-time neural style transfer from a training image to another randomly chosen image in the mini-batch. For ViTs, StyleAug outperforms other state-of-the-art augmentations in accuracy, robustness, transfer learning, and shape bias. We hope that future research will continue to develop augmentations and training strategies that work well for vision transformers, even if they might not benefit the previously dominant CNN architecture.

\clearpage
\bibliography{iclr2020_conference}
\bibliographystyle{iclr2020_conference}

\clearpage
\appendix
\section{Appendix}

\renewcommand{\figurename}{Supplementary Figure}
\setcounter{figure}{0}  

\subsection{Augmentation examples}

Here, we provide examples images of all augmentations that we tested (Supp. Fig. \ref{appendix_fig:aug_examples}). In the examples, "Orig" represents the image after Inception-style preprocessing (random crop with a large scale (0.5,1.0) and horizontal flip). "Aug1" represents the augmentation we used for training when no JSD loss was used. When a JSD loss was used, we trained with the 3 versions of the displayed image. Where applicable, the PyTorch torchvision transform used for augmentation is described in the caption.

\begin{figure}[ht]
    \captionsetup{labelformat=simple}
    
    \begin{subfigure}[a]{1.0\textwidth}
    \begin{center}
        \includegraphics[width=0.9\textwidth]{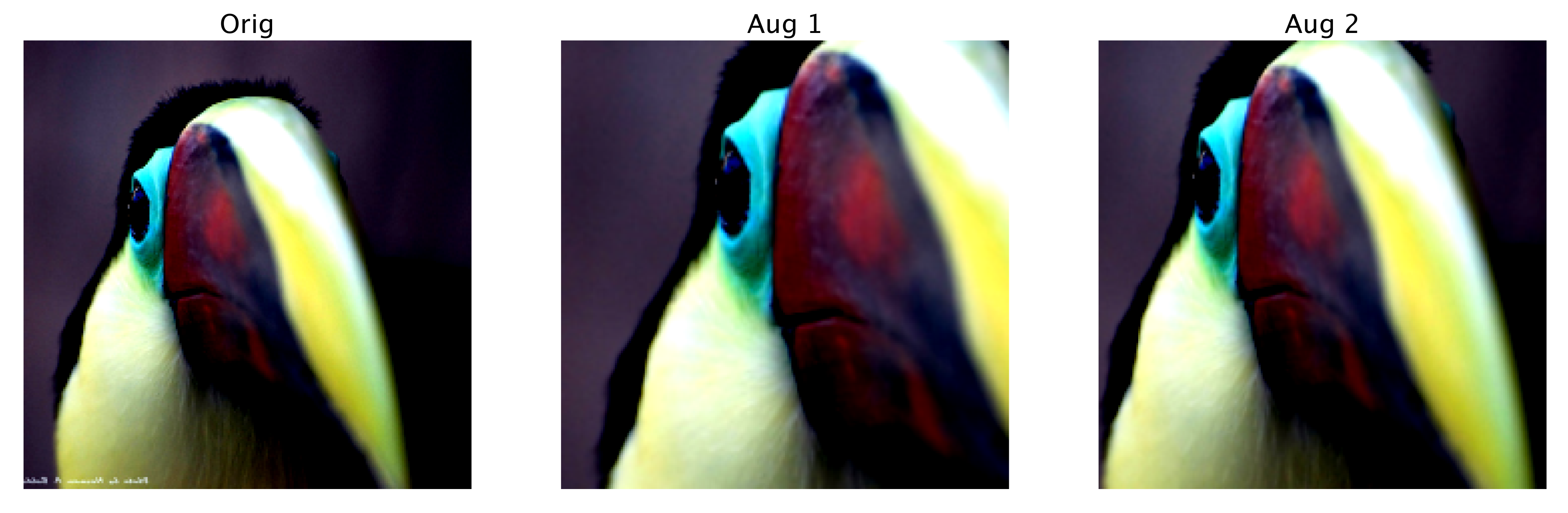}    
    \end{center}
    \caption[Crop augmentation]{Cropping augmentation. RandomResizedCrop(scale=(0.25,1.0)).}
    \label{appendix_fig:crop}
    \end{subfigure}
    
    \begin{subfigure}[b]{1.0\textwidth}
    \begin{center}
        \includegraphics[width=0.9\textwidth]{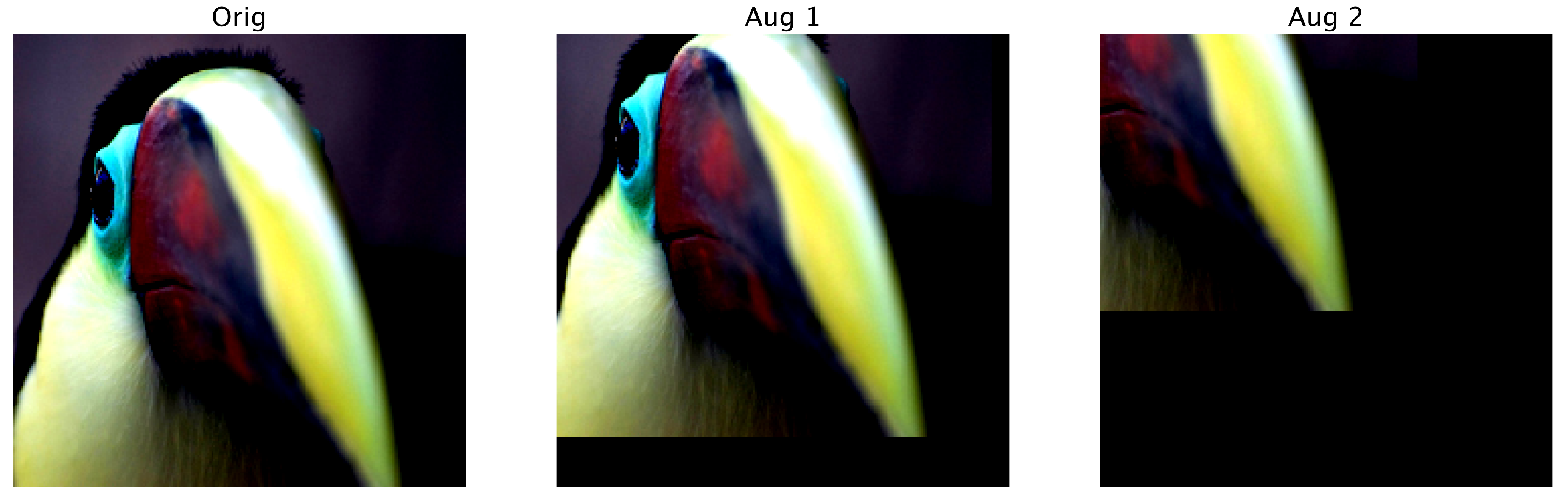}
    \end{center}
    \caption[Translation augmentation]{\textbf{Translate augmentation.} RandomAffine(rotate=0,translate=(0.5,0.5),scale=None,shear=None).}
    \label{appendix_fig:translate}
    \end{subfigure}
    
    \begin{subfigure}[b]{1.0\textwidth}
    \begin{center}
        \includegraphics[width=0.9\textwidth]{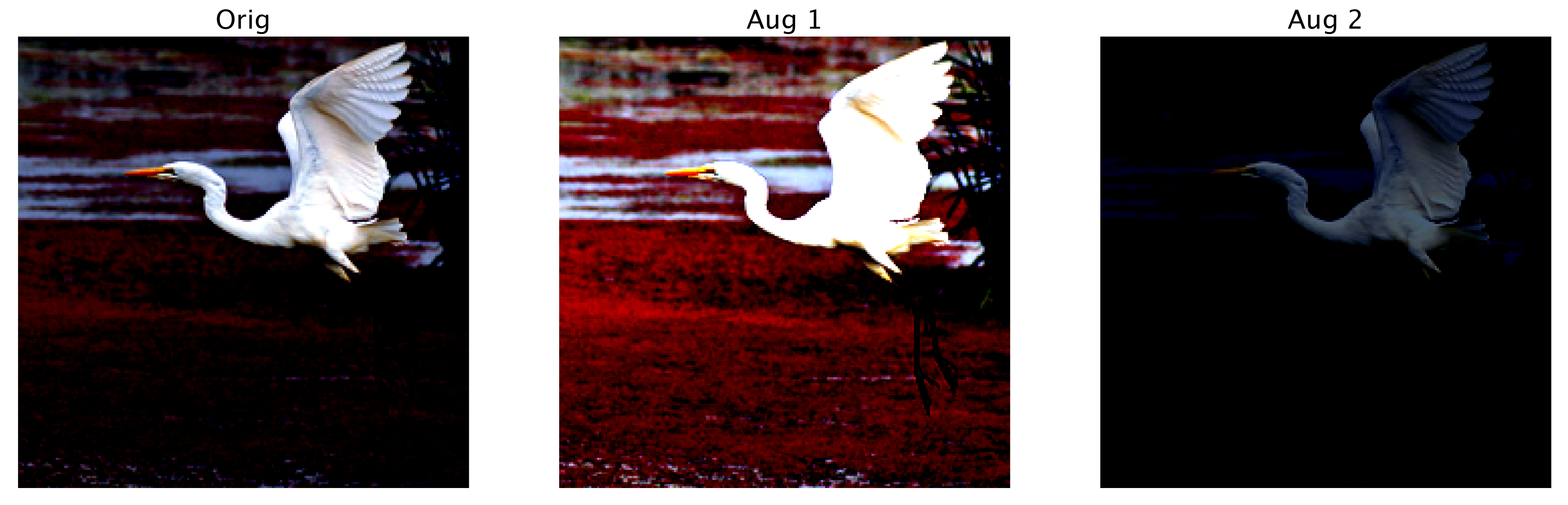}
    \end{center}
    \caption[Color augmentation]{\textbf{Color augmentation.} ColorJitter(brightness=0.4,contrast=0.4,saturation=0.2,hue=0.1).}
    \label{appendix_fig:color}
    \end{subfigure}
    
    \begin{subfigure}[b]{1.0\textwidth}
    \begin{center}
        \includegraphics[width=0.9\textwidth]{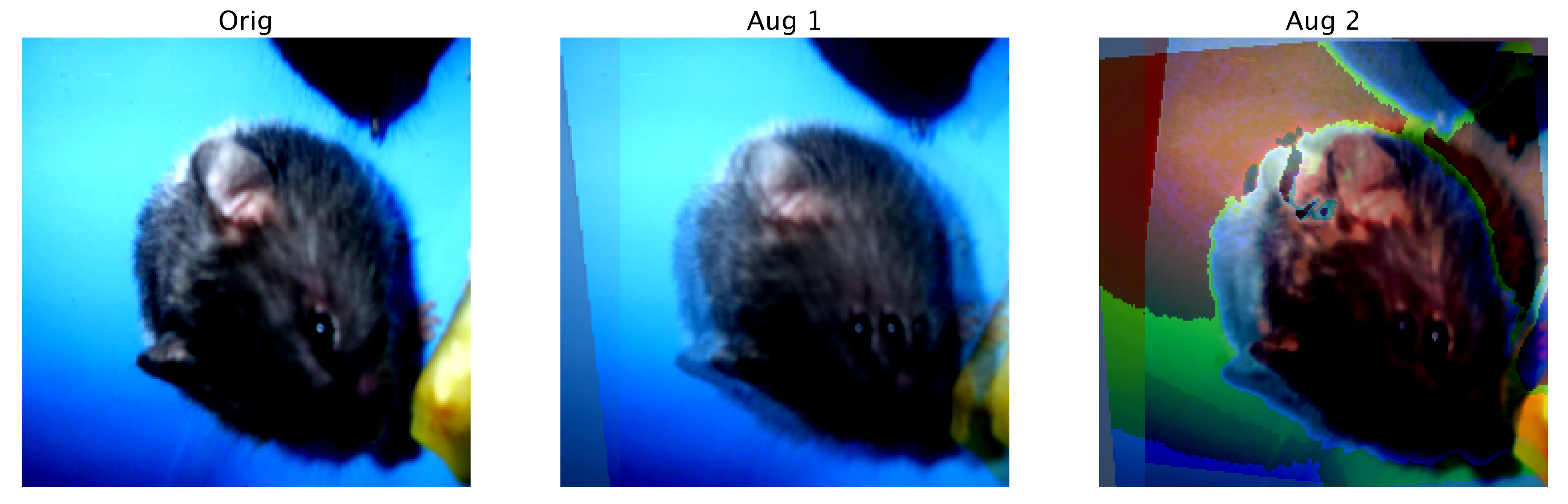}    
    \end{center}
    \caption[AugMix augmentation]{\textbf{AugMix} \citep{hendrycks_augmix_2020}. We used the PyTorch image models (timm; \citep{rw2019timm}) implementation of AugMix. As in the original AugMix work, the set of transformations used were mutually exclusive with the transfomations present in the Imagenet-C dataset \citep{hendrycks_augmix_2020}.}
    \label{appendix_fig:augmix}
    \end{subfigure}

\end{figure}

\begin{figure}[ht]\ContinuedFloat
    \captionsetup{labelformat=simple}
    \centering
    
    \begin{subfigure}[b]{1.0\textwidth}
    \begin{center}
        \includegraphics[width=0.9\textwidth]{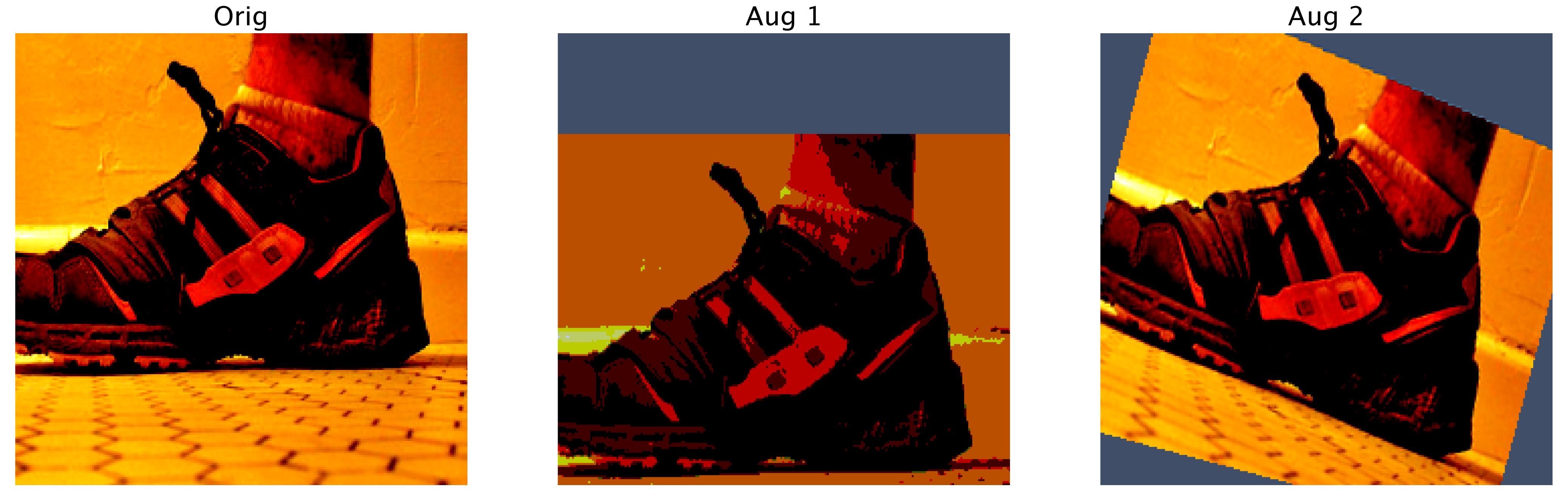}
    \end{center}
    \caption[RandAugment augmentation]{\textbf{RandAugment} \citep{cubuk_randaugment_2019}. We used the PyTorch image models (timm; \citep{rw2019timm}) implementation of RandAugment. We again removed any transformations that overlapped with corruptions in the ImageNet-C dataset \citep{hendrycks_augmix_2020}.}
    \label{appendix_fig:randaug}
    \end{subfigure}
    
    \begin{subfigure}[b]{1.0\textwidth}
    \begin{center}
        \includegraphics[width=0.9\textwidth]{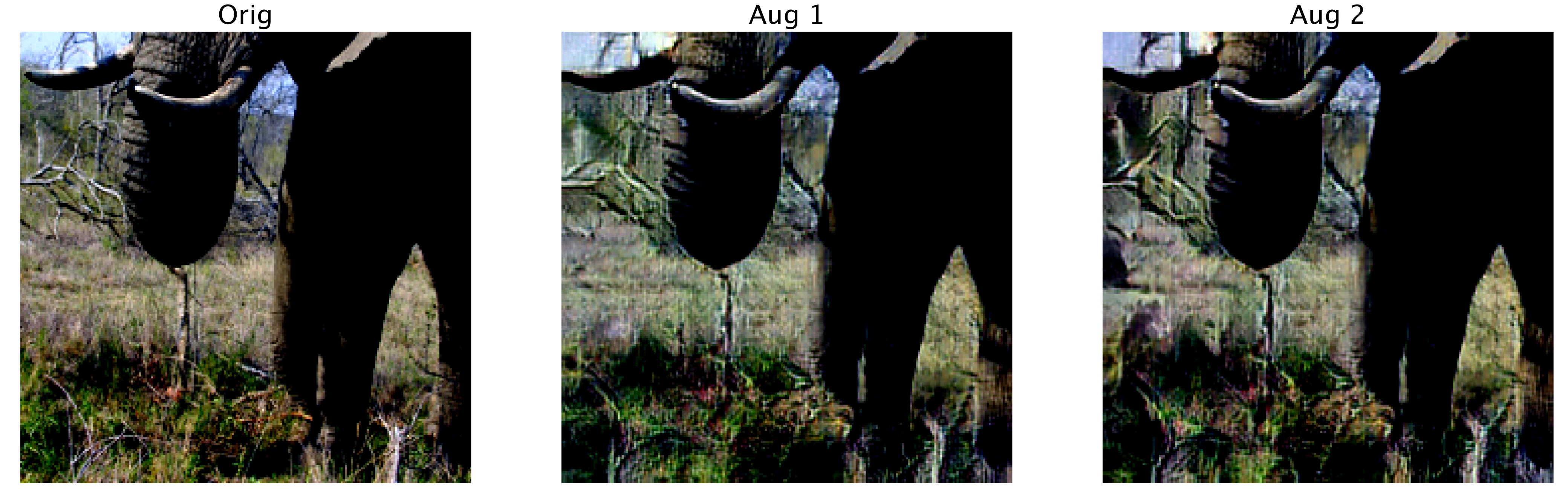}
    \end{center}
    \caption[Neurofovea augmentation]{\textbf{Neurofovea}. We adapted the Neurofovea transformation described by \citet{deza_towards_2018}. For the foveation step, we simply mixed the original image and style-transferred noise image using a weighted mask. The original image received larger weights for points closer to the foveation point with exponentially decaying weights for farther pixels, with a minimum weight of 0.25. The style-transferred noise image received weights that were 1 minus the weights for the original image. \citet{deza_towards_2018} used a more perceptually accurate transformation. However, the computational cost of that approach made it infeasible as an augmentation strategy.}
    \label{appendix_fig:nf}
    \end{subfigure}
    
    \begin{subfigure}[b]{1.0\textwidth}
    \begin{center}
        \includegraphics[width=0.9\textwidth]{final_figures/aug_examples/texture.pdf}
    \end{center}
    \caption[StyleAug augmentation]{\textbf{StyleAug (ours)}. The augmentation is detailed in section 4 of the main text.}
    \label{appendix_fig:style}
    \end{subfigure}
    
    \begin{subfigure}[b]{1.0\textwidth}
    \begin{center}
        \includegraphics[width=0.9\textwidth]{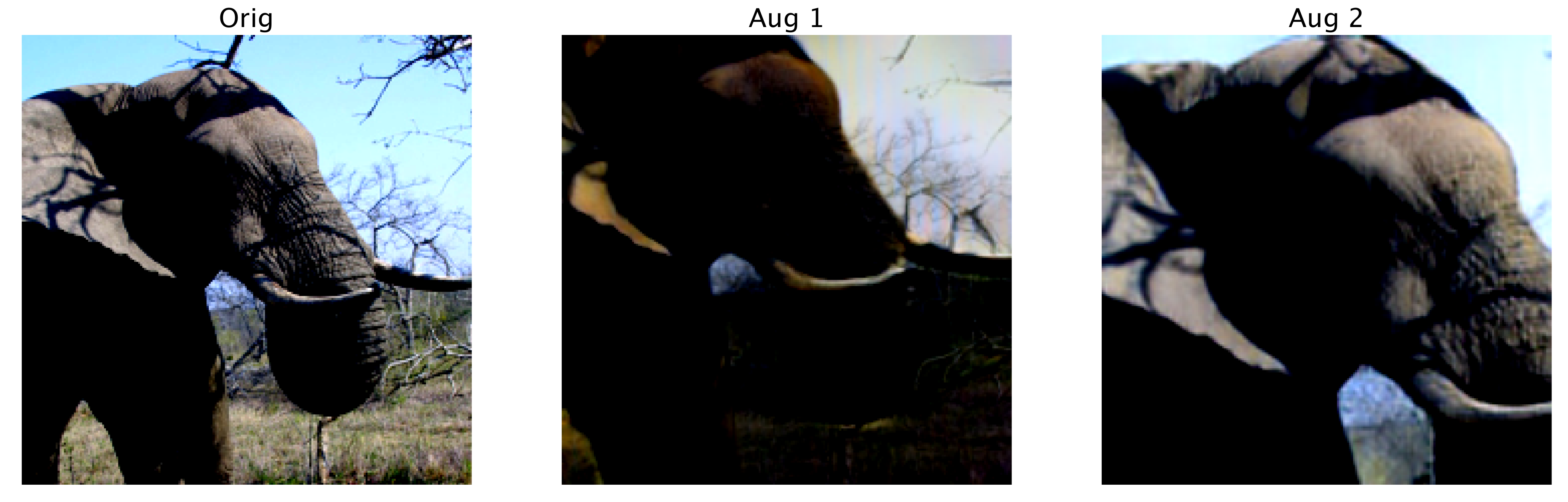}
    \end{center}
    \caption[StyleAug and crop]{\textbf{StyleAug and crop (ours)}. As above in panel (g), with the additional step of RandomResizedCrop(scale=(0.25,1.0)).}
    \label{appendix_fig:styleCrop}
    \end{subfigure}
    
\caption[Augmentation examples]{Example augmentations. "Orig" shows an example image after Inception-style preprocessing. "Aug 1" and "Aug 2" show augmentations applied to "Orig".}
\label{appendix_fig:aug_examples}
\end{figure}

\subsection{Cue-conflict examples}

\begin{figure}[t]
    \captionsetup{labelformat=simple}
    \centering
    \includegraphics[width=.5\textwidth]{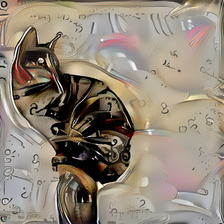}
    \caption[Example cue-conflict image.]{\textbf{Example of a shape vs. texture cue-conflict image \citep{geirhos_imagenet-trained_2018}.} The shape of the object is a cat and the texture of the image is clocks. The displayed image was generated by \citet{geirhos_imagenet-trained_2018}.}
    \label{appendix_fig:cueConflict}
\end{figure}

Example image of the cue conflict experiment dataset generated from \citep{geirhos_imagenet-trained_2018} using style transfer. Each image has two "correct" labels, one relating to dominant shape of the image, and one relating to the dominant texture of an image. In the example image in Supp. Fig. \ref{appendix_fig:cueConflict}, the shape of the object is a cat and the texture of the image is clocks.

\end{document}